# Adaptive Recruitment Resource Allocation to Improve Cohort Representativeness in Participatory Biomedical Datasets

Victor A. Borza, B.E.[1], Andrew Estornell, Ph.D.[2,3], Ellen Wright Clayton, M.D., J.D.[1], Chien-Ju Ho, Ph.D.[3], Russell L. Rothman, M.D., M.P.P.[1], Yevgeniy Vorobeychik, Ph.D.[3], Bradley A. Malin, Ph.D.[1]

[1]Vanderbilt University, Nashville, TN; [2]ByteDance Research, San Jose, CA; [3]Washington University in St. Louis, St. Louis, MO

**Abstract**

*Large participatory biomedical studies – studies that recruit individuals to join a dataset – are gaining popularity and investment, especially for analysis by modern AI methods. Because they purposively recruit participants, these studies are uniquely able to address a lack of historical representation, an issue that has affected many biomedical datasets. In this work, we define representativeness as the similarity to a target population distribution of a set of attributes and our goal is to mirror the U.S. population across distributions of age, gender, race, and ethnicity. Many participatory studies recruit at several institutions, so we introduce a computational approach to adaptively allocate recruitment resources among sites to improve representativeness. In simulated recruitment of 10,000-participant cohorts from medical centers in the STAR Clinical Research Network, we show that our approach yields a more representative cohort than existing baselines. Thus, we highlight the value of computational modeling in guiding recruitment efforts.*

**Introduction**

Spurred by recent innovations in artificial intelligence (AI) and machine learning (ML), large biomedical datasets are providing revolutionary insights into health and healthcare. Many of these datasets involve secondary uses of existing data, such as electronic health records (EHRs) collected by academic medical centers in the context of patient care.[1] In contrast to these secondary use datasets, participatory datasets are comprised of purposively recruited participants who choose to join and share their data. There has been recent interest, as well as billions of dollars of investment from funding agencies such as the National Institutes of Health, in multi-site participatory datasets including the Bridge2AI Voice project,[2] AI-READi,[3] and the *All of Us* Research Program.[4] Participatory datasets yield several benefits over data originating from non-recruited settings. Notably, programs like *All of Us* include participants as partners,[5] which may guide research study design and improve inclusion and communication between researchers and participants.[6] Moreover, participatory studies can adapt their recruitment practices to align their cohort with certain goals, like being representative of a population or improving recruitment of groups historically underrepresented in research.[3,7] While recognizing that several definitions of representativeness have been proposed,[8–11] in this work we rely on a definition common in the literature that a *representative* cohort is similar to a target population when considering some attributes of interest.

Diverse representation is critical for biomedical research for several reasons.[12,13] First, per the 1979 Belmont Report, underrepresentation of population subgroups may reflect institutionalized biases and conflict with the Principle of Justice.[14,15] Second, representation supports perceptions of legitimacy.[16] Conversely, lack of representation may undermine trust in the research enterprise.[17] Third, representation bias can skew the associations that researchers may identify.[18] Moreover, the efficacy of predictive models in specific subgroups can vary widely based on the representation of those subgroups in training datasets, which has contributed to multiple forms of non-trivial bias in biomedical and non-biomedical settings.[19–21] Fourth, lack of representation may compromise the generalizability of research to underrepresented groups.[17] Often, these underrepresented groups are also marginalized. Thus, lack of representation limits innovation in, as well as access to, effective medical interventions, and costs hundreds of billions of dollars via perpetuated health disparities and inequities.[17] Despite these numerous harms, biomedical research cohorts across fields of medicine have struggled to sufficiently represent the entire population they intend to study.[22–27] Within informatics, demographic data are incompletely reported in many studies applying ML models to EHR data, and when model developers report demographics, they often reveal unrepresentative training datasets.[27] Thus, there is an urgent need for methods to improve representation in participatory biomedical datasets, a need echoed by the U.S. White House.[28]

There have been several investigations into improving representation in datasets, though few have been directly applicable to recruitment in multi-site participatory datasets. When subjects or records can be selected according to the attributes that define representation, or any other use-specific constraints like quotas, more representative cohorts can be constructed.[29–31] Other studies have described processes for sampling records to improve algorithmic fairness (i.e., decrease disparate predictions or performance between groups) when a specific downstream task is known *a priori*.[32,33] However, these methods assume that subjects or records may be selected according to predefined attributes of interest, and some assume that the study team already knows their specific analysis or task. In prospective recruitment with participatory studies, especially multi-purpose cohorts like *All of Us* or the U.K. Biobank, these strategies may not be feasible or cost-effective. Consequently, we conceptualize a recruitment policy as a specific allocation of resources among recruitment sites. The recruitment policy is enacted in simulation as stochastic draws from each site's demographic distribution. This conceptualization lends itself well to a multi-armed bandit (MAB) model, a well-known framework commonly used for reinforcement learning.[34] Notably, Nargesian et al. proposes a similar method for achieving a desired cohort attribute distribution from multiple data sources, but their method is not adapted for recruitment and may require one to discard collected data, a highly undesirable action in medical settings.[35]

We model multi-site recruitment of research cohorts as an MAB problem of allocating recruitment resources – and, thus, expected recruitments – among sites to yield a representative cohort. We adapt the MAB problem to recruitment in two key ways: 1) recruiting multiple individuals at each time step and allowing multiple sites to be simultaneously recruited from, and 2) modeling causal biases that induce a site's distribution to drift non-randomly, either over time or in response to recruitment. We hypothesize that adaptive recruitment policies will generate more representative cohorts than both naïve policies – which disregard demographics information – and existing MAB baselines.[34,36] We assess this hypothesis using demographic data from an existing clinical research network that we use as a proxy for sites recruiting individuals into a hypothetical participatory biomedical dataset.

**Methods**

*Data Sources and Processing:* To simulate recruitment of a multi-site participatory biomedical dataset, we use summary demographic data from the Stakeholders, Technology, and Research (STAR) Clinical Research Network (CRN). STAR is a network of nine health systems that together comprise one of the eight CRNs funded by the Patient Centered Outcomes Research Institute (PCORI).[37–39] As of the most recent public data release (last updated November 16, 2022), the STAR CRN reports demographic information on age (in 4 bins), gender, race, and ethnicity, for their sites: 1) Vanderbilt University Medical Center (VUMC), 2) Vanderbilt Health Affiliated Network (VHAN), 3) Meharry Medical College (MHRY), 4) Mayo Clinic (MAYO), 5) University of North Carolina (UNC), 6) Duke University (DUKE), 7) Medical University of South Carolina (MUSC), 8) Spartanburg Regional Healthcare System (SRHS), and 9) Wake Forest (WAFO). In total, the nine sites comprise 18.5M patients as of their last public reporting updates, ranging from July to September 2022. We recognize that some patients may be included in multiple sites because of geographic overlap between the sites' areas of service.

We exclude all records with values of 'No Information', 'Unknown', 'Ambiguous', 'Refuse to answer', or 'Other' in any demographic fields for harmonizing with U.S. Census data, yielding 14.5M (78.4%) records for analysis. Additionally, we note that only univariate demographic distributions are available (e.g., patients by age *or* by gender but not by age *and* gender), such that we assume attribute independence to construct a joint multivariate demographic distribution at each site. To best estimate the true population of the U.S., we rely on the July 2022 U.S. Census Bureau population estimates by age (in 5-year bins), sex, race, and ethnicity from the American Community Survey (table CC-EST2022-ALL), based on the 2020 decennial Census.[40,41] To harmonize with STAR CRN demographics data, we focus on individuals identifying with one race only, yielding a U.S. population of 323.2M. We distribute the 15-19 years age bin in Census data proportionally, with 60% of the population being allocated to the 0-17 years bin and 40% to the 18-45 years age bin.

*Quantifying Representation*: We quantify representation as the similarity between a cohort and a target population across four measured attributes of interest: age, gender, race, and ethnicity. To assess representativeness, we measure its inverse, a statistical distance between the cohort and target population. Lower distances indicate greater representativeness, with zero distance indicating perfect reflection of target population demographics. Many such distance measures exist, yet it is unknown which one is appropriate. As such, we considered three in our analyses. First, we utilized the Distance Summary measure from Borza et al.,[31] which corresponds to a normed sum of univariate Jensen-Shannon distances across the marginal distributions of attributes. Second, we utilized the sum of univariate Kullback-Leibler divergences (KLDs) across the marginal distributions of attributes. Third, we consider the single multivariate KLD across the joint distribution of all attributes.[42]

**Algorithm 1.** Adaptive recruitment strategies for improving representation in simulated recruitment

    **Data:** sites $S_1 \ldots S_n$ with response distributions $D \in \mathbb{M}_{n \times m}$ (matrix of $n$ sites with $m$ groups), target population distribution $P \in \mathbb{R}^m$, total iterations $T$, final cohort size $N$, distance function $F$
    **Result:** Cohort $C$

```
1   C ← ∅
2   initialize Dirichlet parameters α;                                    /* based on prior knowledge */
3   for t = 1, …, T do
4   |   ρ ∈ ℝ≥0ⁿ ← 0;                                                    /* initialize resource allocation vector */
5   |   for j = 1, …, n do
6   |   |   D̂_j[t] ~ Dir(α_{j,1}, …, α_{j,m});                          /* estimate response distributions at iteration t */
7   |   end
8   |   if "Thompson" then ρ*_{j*} = 1 where j* ← arg min_j 𝔼[F(D̂_j[t] + C, P)];      /* single best site */
9   |   else then ρ* ← arg min_ρ 𝔼[F(ρD̂[t] + C, P)] subject to ∑ρ = 1;   /* best combination of sites */
10  |   r_1, …, r_n = round(ρ*_{1…n} × N/T);          /* force integer recruitment counts and sum to N/T */
11  |   for j = 1, …, n do
12  |   |   c ← ∅;                                                        /* set of recruitments from site j */
13  |   |   for i = 1, …, r_j do
14  |   |   |   c += s ~ Cat(D_j[t]);                                    /* recruit an individual from site j */
15  |   |   end
16  |   |   C, α_i += c;                              /* Add recruited individuals to cohort and update prior */
17  |   end
18  end
```

**Table 1.** Overview of experimental parameters and terminology

| Response Distribution | Model | Prior Information |
|---|---|---|
| *Static:* Site demographics and response distributions are the same | *Naïve:* Recruitment policy is chosen without respect to site demographics | *Uninformed:* Jeffreys prior is used to estimate site demographics |
| *Shifting:* Response distributions change with each time step | *Adaptive:* Recruitment policy is actively changed based on site demographics | *Fully informed:* Actual site demographics are used as a prior |
| *Causal Bias:* Response distributions change when a site is recruited from | | *Empiric:* Prior is established through pre-experiment sampling |

*Recruitment Simulation*: We formalize the goal of achieving representativeness through recruitment as an MAB problem.[34,43] The recruitment timeline is split into a sequence of recruiting iterations according to the frequency of policy updates. We model a 5-year recruitment timeline of 10,000 participants with quarterly policy updates as a 20-step process with 500 recruitments per iteration, where the resources to achieve these 500 recruitments may be allocated among sites. We acknowledge that various factors may influence the participation rates of different groups in actual recruitment. Thus, we consider the product of a site's demographic distribution with its subgroup-specific participation rates to yield a *response distribution*, which corresponds to the expected distribution of successful recruitments for each site. For example, a subgroup with a higher participation rate will have a higher probability density in the response distribution than a subgroup with a lower participation rate, even if the two groups comprise equal proportions of the demographic distribution. In the basic recruitment case, we assume that all groups have equal participation rates that do not change through recruitment, such that we may use site demographic distributions as the response distributions. To simulate changes in the response distribution (Table 1), we systematically modify the response distribution as a function of each subgroup's proportion of the demographic distribution, such that site-majority groups all become either more or less likely to respond. When these changes occur over time, independent of recruitment policies, we call the changes distribution shifts (Eq. 1). When these changes occur in response to recruitment, they are termed causal bias (Eq. 2).

$$D_{j,i}[t+1] = D_{j,i}[t]^\lambda / \sum_{i=1}^m D_{j,i}[t]^\lambda \quad \text{Distribution Shift (Eq. 1)}$$

$$D_{j,i}[t+1] = D_{j,i}[t]^{(1+\rho_j(\kappa-1))} / \sum_{i=1}^m D_{j,i}[t]^{(1+\rho_j(\kappa-1))} \quad \text{Causal Bias (Eq. 2)}$$

where $\lambda$ is the distribution shift factor, $\kappa$ is the causal bias factor, and $D$ and $\rho$ come from Algorithm 1. A value of 1 for either of these factors recovers the no-bias case; a value greater than 1 indicates site-majority groups are more likely to respond (accentuating the shape of the site's underlying demographic distribution); and a value less than 1 indicates site-minority groups are more likely to respond (blunting the shape of the site's demographic distribution).

*Policy Optimization*: In each iteration, recruiters have some knowledge of each site's response distribution, modeled as a categorical distribution drawn from a Dirichlet conjugate prior distribution. In the first iteration, we initialize *a priori* knowledge of site response distributions in one of three ways (Table 1): 1) an *uninformed* (Jeffreys) *prior* with concentration parameters α = 0.5, 2) *empirically* through pre-simulation sampling from each site, or 3) a *fully informed prior* using actual site demographic distributions as concentration parameters. By using these priors, we cover the range of possible starting information that may be available to researchers starting recruitment. In subsequent iterations, simulated recruitments are used to update the Dirichlet conjugate prior and represent increased knowledge of the site response distributions. Recruiters may use this prior knowledge to determine an optimal allocation of recruitment resources among the sites, a process we show in Algorithm 1.

We analyze two broad classes of recruitment strategies (Table 1): 1) *naïve* strategies, which disregard knowledge of site demographics during policy updates and 2) *adaptive* strategies, which use this knowledge to update recruitment policies. The three naïve strategies we analyze are: 1) "Random Site," which allocates all resources to one random site each iteration, 2) "Uniform," which allocates equal resources to each site at each iteration, and 3) "Informed Static," which simulates 1,000 samples from each site to develop an empiric prior and identifies a policy to optimize representativeness which is not updated over iterations. Although not involving recruitment, some multi-site genomic analyses roughly apply a uniform distribution of sampling across sites.[44] Moreover, the informed static policy represents a partially informed, but fixed, resource allocation (e.g., fixed grant funding). To evaluate a baseline adaptive strategy, we use Thompson sampling (Algorithm 1, line 8),[34,36] which allocates all recruitment resources to the site which is expected to reduce a given distance function the most. Finally, our proposed adaptive strategy (Algorithm 1, line 9) distributes resources among sites in a locally optimal manner for a given distance function. We provide the results for two strategies with different distance functions: 1) multivariate KLD and 2) Distance Summary. After the recruitment policy – defined as the number of participants recruited at each site during each step – is set, a stochastic recruitment process is applied. We model recruitment as random draws from each site's response distribution, incorporating both underlying demographics and group-specific participation rates.

*Computation and Statistics:* All experiments were conducted in Python 3.11. Due to stochasticity in our recruitment simulation, 100 experiments were conducted for each unique set of parameters. 95% Bayesian credible intervals for the means (across experiments) of the three analyzed distance measures at each iteration were calculated using a non-informative prior.[45] If the posterior distributions within two credible intervals were normally distributed, then non-overlapping intervals roughly align with rejecting a frequentist null hypothesis at α = 0.01.[46]

**Results**

**Table 2.** STAR CRN site characteristics and demographic ratios relative to the U.S. Census. Positive (blue) values indicate overrepresentation of a group compared to Census counts while negative (red) values indicate underrepresentation. Abbreviations not in text: *MKLD* [multivariate KLD] and *NH/L* [non-Hispanic/Latino].

| Site: | US Census | VUMC | VHAN | MHRY | MAYO | UNC | DUKE | MUSC | SRHS | WAFO |
|---|---|---|---|---|---|---|---|---|---|---|
| State | All | TN | TN | TN | MN/AZ/FL | NC | NC | SC | SC | NC |
| Count | 323.18M | 1.978M | 316.3K | 311.7K | 2.808M | 3.055M | 2.728M | 1.305M | 521.7K | 1.513M |
| Missing | N/A | 25.26% | 9.05% | 20.50% | 9.01% | 26.42% | 27.80% | 12.47% | 19.64% | 22.62% |
| MKLD | N/A | 0.1523 | 0.2170 | 0.4875 | 0.2477 | 0.1314 | 0.1251 | 0.2583 | 0.1556 | 0.1199 |
| **Age** | | | | | | | | | | |
| 0-17 | 21.07% | 1.237 | -1.250 | -1.944 | -2.218 | -1.283 | -1.275 | -1.111 | -1.275 | -1.290 |
| 18-44 | 36.10% | -1.082 | -1.067 | 1.180 | -1.205 | -1.061 | -1.059 | -1.123 | -1.028 | -1.130 |
| 45-64 | 25.13% | -1.092 | 1.152 | 1.291 | 1.102 | 1.031 | 1.037 | -1.067 | 1.037 | -1.015 |
| 65+ | 17.69% | -1.008 | 1.151 | -1.252 | 1.857 | 1.335 | 1.319 | 1.431 | 1.260 | 1.523 |
| **Gender** | | | | | | | | | | |
| Female | 50.41% | 1.068 | 1.119 | -1.036 | 1.030 | 1.091 | 1.095 | 1.081 | 1.042 | 1.069 |
| Male | 49.59% | -1.075 | -1.138 | 1.035 | -1.031 | -1.102 | -1.106 | -1.090 | -1.045 | -1.075 |
| **Race** | | | | | | | | | | |
| AI/AN | 1.35% | -6.374 | -26.52 | -3.045 | -2.339 | 1.539 | -2.385 | -7.442 | -5.604 | -1.894 |
| Asian | 6.48% | -3.078 | -3.501 | -8.111 | -2.478 | -2.691 | -2.547 | -5.912 | -4.460 | -3.211 |
| Black | 14.05% | -1.051 | -1.984 | 3.586 | -3.115 | 1.695 | 1.693 | 2.295 | 1.616 | 1.363 |
| NH/PI | 0.27% | -1334.7 | -7.210 | -1.500 | -1.558 | -1.863 | -2.564 | -2.700 | -1.132 | -2.087 |
| White | 77.85% | 1.083 | 1.169 | -1.615 | 1.183 | -1.088 | -1.066 | -1.173 | -1.033 | 1.002 |
| **Ethnicity** | | | | | | | | | | |
| H/L | 19.05% | -3.566 | -5.670 | -1.376 | -3.423 | -2.039 | -2.037 | -4.488 | -3.219 | -2.021 |
| NH/L | 80.95% | 1.169 | 1.194 | 1.064 | 1.167 | 1.120 | 1.120 | 1.183 | 1.162 | 1.119 |

*Demographics of STAR CRN Sites:* Table 2 presents the demographic distributions from the U.S. Census, along with the ratios of each population group at the nine STAR CRN sites to the Census. Positive ratios indicate a group is overrepresented compared to the Census while negative ratios indicate underrepresentation. General trends include underrepresentation of younger individuals, males, individuals identifying with a race of 1) American Indian / Alaska Native (AI/AN), 2) Asian, 3) Native Hawaiian / Pacific Islander (NH/PI), and/or Hispanic / Latino (H/L) ethnicity.

*Recruiting a More Representative Cohort:* Figure 1 shows that, in our baseline cases with static response distributions, adaptive recruitment strategies outperform naïve recruitment strategies at generating a representative cohort. The results were qualitatively similar for the three distance measures analyzed, such that we only present results showing multivariate KLD. As expected, the adaptive recruitment strategy that minimizes multivariate KLD yielded the most representative cohorts on average, using this measure. While the differences in KLD between algorithms may only represent a 20% difference or less, small relative differences understate major impacts on cohort demographics. For instance, Figure 2 highlights the impact of adaptive recruitment strategies on the demographics of the final cohort by comparing the naïve uniform baseline (arrow tail) to our fully informed adaptive strategy (arrow head). The adaptive strategy reduces the underrepresentation of AI/AN and Asian subgroups but can accentuate underrepresentation of some NH/PI subgroups.

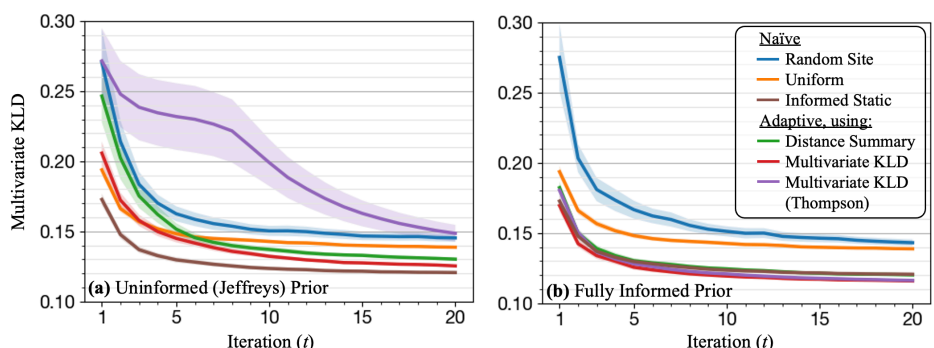

**Figure 1.** A comparison of cohort representativeness (where lower KLD is more representative) for naïve and adaptive recruitment strategies over time, using uninformed **(a)** and fully informed **(b)** prior knowledge. The shaded areas indicate a 95% Bayesian credible interval of mean KLD values.

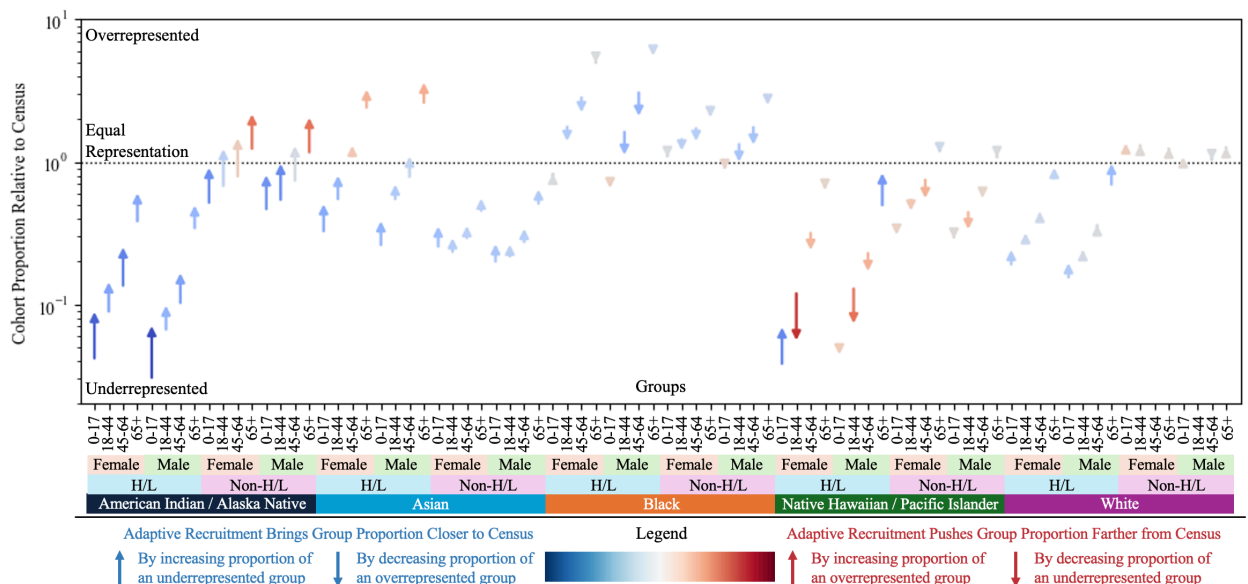

**Figure 2.** Changes in cohort subgroup proportions from a naïve uniform policy (arrow tail) to an adaptive, fully informed policy optimizing for MKLD (arrow head). Blue arrows indicate the adaptive policy brings the subgroup proportion closer to its Census levels (dotted line) while red arrows indicate the proportion departs from Census levels.

Incorporating prior knowledge about site-specific demographic distributions allowed all adaptive sampling methods to recruit more representative cohorts. The MKLD-based adaptive strategy decreased its final KLD from 0.1257 [95%

CI: 0.1247, 0.1266] to 0.1157 [95% CI: 0.1149, 0.1164] (Figure 1), surpassing both the site with the lowest KLD (Wake Forest, 0.1199) and the informed static strategy (0.1206 [95% CI: 0.1193, 0.1218]). The informed static strategy, which uses pre-simulation sampling to establish a prior, represents partially informed sampling. Expectedly, it performs better than uninformed recruitment and worse than fully informed recruitment. The impact of prior knowledge is most noticeable on the Thompson sampling strategy, which transforms from performing worse than naïve baselines to becoming statistically indistinguishable from our best-performing strategy. Due to the high variance of the Jeffreys prior, the uninformed Thompson sampler often allocates resources to the same site during the first several iterations, a phenomenon evidenced by bands of constant resource allocation in the early iterations in Figure 3c. Since the distributed adaptive recruitment strategy may sample from multiple sites at each iteration, this strategy explores more efficiently in the initial steps than the Thompson sampler. After only five iterations, our uninformed adaptive strategy learns sufficient information about site demographics to outperform the naïve baselines. After ten iterations, the adaptive policies begin to resemble those of their fully informed counterparts (Figure 3a-b).

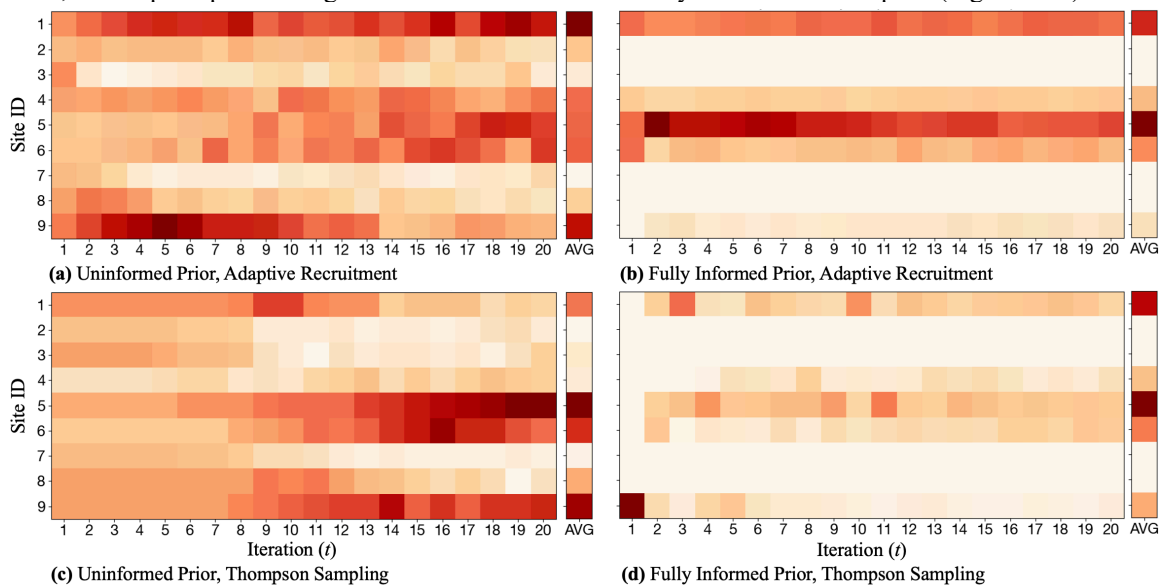

**Figure 3.** Mean (across experiments) recruitment policies at each iteration and averaged across all iterations (AVG) for our distributed adaptive recruitment policy **(a, b)** compared to a Thompson sampler **(c, d)**, both with **(d, b)** and without **(a, c)** prior knowledge of site response distributions. Darker colors indicate higher recruitment density.

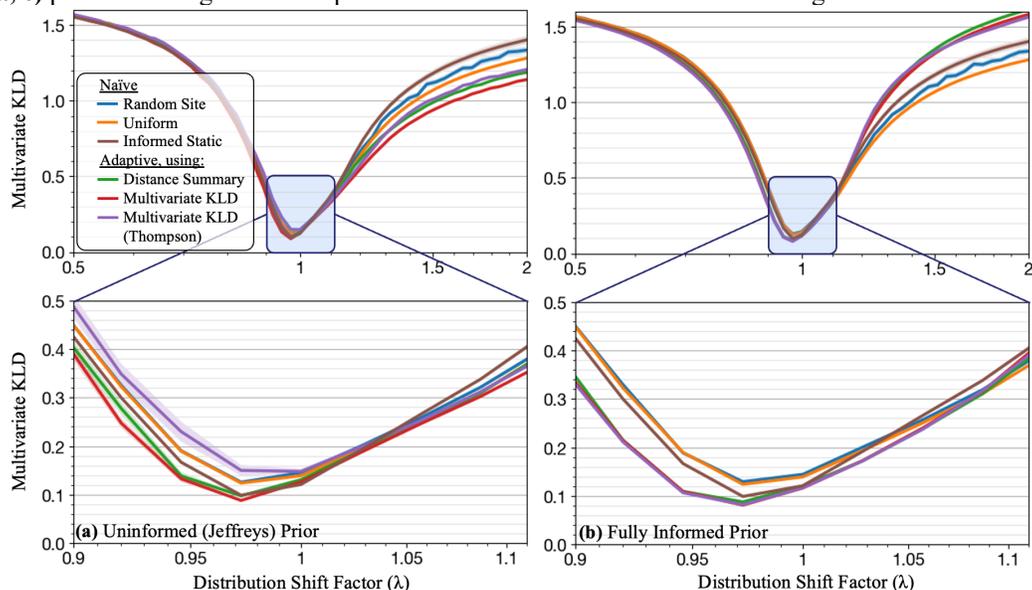

**Figure 4.** The final cohorts of uninformed adaptive strategies are more representative than those of naïve strategies across distribution shift factors λ **(a)** while the fully informed adaptive strategies suffer from strong, yet incorrect,

priors and show lower representativeness when λ is much greater than 1 **(b)**. The shaded regions around the lines indicate a 95% Bayesian CI and the blue boxes in the top row correspond to the magnified areas in the bottom row.

*Varying the Response Distribution:* Real-world recruitment scenarios often do not follow the strict arm independence and invariance conditions of traditional MAB analyses. To simulate situations where the response distribution changes either over time or in response to recruitment, we investigated a range of distribution shift and causal bias factors and analyzed their effect on final cohort KLD (Figures 4, 5). Shifts in the response distribution challenged our adaptive recruitment strategies because both prior knowledge and knowledge gained through recruitment become less accurate over time. In the distributional shift analysis, uninformed adaptive models performed significantly better than their fully informed counterparts for distribution shift factors greater than 1 (Figure 4). Across the range of tested distribution shift factors, our uninformed adaptive strategy consistently recruited more representative cohorts than the naïve baselines or Thompson sampler. Expanding on changes to the response distribution, causal bias presents a yet more challenging situation: recruiting at a site may induce a bias that makes the site sub-optimal. Thus, with both uninformed and fully informed adaptive strategies, there is a range of causal bias factors where adaptive strategies outperform naïve baselines (Figure 5). This range is approximately 0.85 – 1.25 for uninformed strategies and 0.8 – 1.05 for fully informed strategies. Thus, causal bias in the response distribution of >5-25% per iteration lead the estimated response distributions to accumulate so much error over time that adaptive recruitment becomes ineffective. Similar to our results for distribution shift factors, uninformed strategies trade off base-case optimality for greater flexibility in the face of changing response distributions.

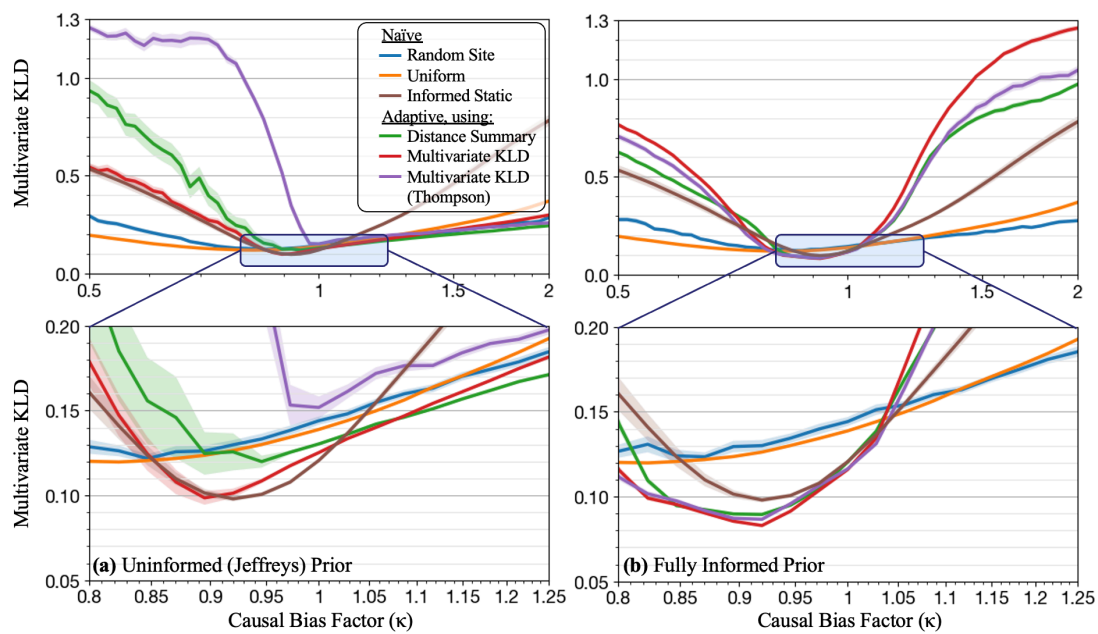

**Figure 5.** The final cohorts of both uninformed **(a)** and fully informed **(b)** adaptive strategies are more representative than those of naïve strategies for a range of causal bias factors κ between 0.85 – 1.25 and 0.8 – 1.05, respectively. The shaded regions around lines indicate 95% Bayesian CI and the blue shaded boxes in the top row indicate areas that are magnified in the bottom row.

In summary, we find that our methodology outperforms both naïve and other adaptive baselines through efficient exploration in a scenario where the total number of budget re-allocations (i.e., iterations) are quite limited. Moreover, we show that our proposed recruitment strategy is robust to shifting distributions of interested participants at each site.

**Discussion and Conclusions**
Our analyses show some notable trends that provide insight into recruitment resource allocation in multi-site projects. Although the uninformed adaptive strategies do not yield the most representative cohorts overall, their benefits over naïve baselines are still notable. Response distributions of the recruitable populations at each site are likely to be unknown before recruitment starts, so a methodology that does not rely on this prior knowledge may be more practical. Moreover, we show uninformed methods to be more robust to time-dependent distributional shifts and recruitment-induced biases than informed methods, strengthening their utility. As we note, uninformed methods sacrifice some best-case optimality to achieve this flexibility. In other words, varying the amount of prior information shifts the strategy's balance between exploration (establishing which sites are best to recruit at) and exploitation (actually recruiting at those sites), a fundamental concept in reinforcement learning and MAB.[34]

Another interesting result in this study is an apparent local optimum effect. Despite site 9's (Wake Forest's) achieving the lowest KLD of all the STAR CRN sites, it receives only a small fraction of recruitment resources under the fully informed adaptive policies. In the uninformed case, the site receives the highest proportion of recruitment resources during the first ten iterations. As the adaptive recruitment model learns more about all the site response distributions, it shifts from a locally optimal recruitment strategy (prioritizing site 9) to a globally optimal strategy (prioritizing sites 4, 5, and 6) that more closely mirrors the strategy of the fully informed model. This result highlights the benefits of multi-site studies and underscores the need for informed recruitment strategies. An informed combination of several less-representative sites may yield a final dataset that is more representative than any single site.

Despite the merits of this research, there are several key limitations to note. First, our measures of representativeness inherently assume a group-based categorization of people and imply these groups are significant. Social concepts like race and ethnicity may drift over time in reporting by the U.S. Census,[47] and should be used with intention and caution in downstream analyses.[48,49] We use group-based analyses to promote inclusion because they are readily available, though we recognize their limitations. Second, we are limited to data and groups that are present in both STAR CRN data and the U.S. Census, leading to the exclusion of individuals identifying with two or more races or missing any demographic attributes, the equating of Census-reported sex with STAR CRN-reported gender, and the inability to assess variables beyond age, gender, race, and ethnicity. Cohorts that are representative across the variables measured may be unrepresentative across unmeasured attributes. Moreover, a significant proportion (9-28%) of each STAR CRN site had missing demographic data that are likely to not be missing completely at random (non-MCAR). Imputation, a strategy used by various federal agencies, including the U.S. Census Bureau,[41] may address missingness but will bias non-MCAR results. Third, because of data availability, we also assumed the independence of demographic attributes to generate joint site distributions. If we applied more accurate joint population distributions, we would expect our methodology to perform even better. While these data limitations could introduce unmeasured biases in our analyses, we would still expect adaptive recruitment resource allocation to improve cohort representativeness over naïve methods.

We recognize some inherent limitations to using the STAR CRN as our example. It is primarily composed of large academic medical centers in urban areas, so it represents fewer individuals living in rural areas and/or experiencing barriers to accessing healthcare. This limitation applies to all studies that primarily recruit participants at academic medical centers. STAR CRN sites are also mostly located in the Southern U.S., so attempting to recruit a nationally representative cohort may be challenging due to geographic variations in demographics. Nevertheless, recruiting a nationally representative cohort remains a reasonable aspiration for participatory studies.

In conclusion, we developed and evaluated a method for improving representation in multi-site participatory biomedical dataset recruitment in this paper. We modified an existing multi-armed bandit framework to make it more relevant to biomedical recruitment and simulate allocation of recruitment resources across the nine-site STAR CRN. Our proposed algorithm recruits cohorts that are more representative than both naïve and adaptive baselines and is flexible to shifts in population and/or response distributions at sites. Moreover, our methodology identified specific and non-obvious recruitment policies to optimize representation, that underscores its potential utility for researchers. There are several areas to consider for future investigation. Weaker informative priors may span a fuller range between optimality and flexibility. Differential site-specific recruitment costs could better reflect real-world differences. Current recruitment practices are not adaptive, so implementation strategies and infrastructure for adaptive recruitment strategies must be developed. Furthermore, the ethical implications and potential unforeseen biases of adaptive recruitment should be studied before these strategies are implemented.

**Acknowledgements**
We thank Dr. Jessica Ancker for constructive review and feedback. Borza is sponsored by NHLBI/NIH grant F30HL168976, while Malin and Clayton are supported by NHGRI/NIH grant U54HG012510. The STAR CRN is supported by PCORI contract RI-CRN-2020-009 and the Vanderbilt Institute for Clinical and Translational Research with funding from NCATS/NIH grant UL1TR002243 and institutional funding.